\documentclass[runningheads]{llncs}
\usepackage[T1]{fontenc}
\usepackage{graphicx}
\usepackage[hidelinks]{hyperref}
\usepackage{booktabs}
\usepackage{multirow}
\usepackage{subcaption}
\usepackage{cite}
\usepackage{xcolor}
\usepackage{graphicx}
\usepackage{amsmath}
\usepackage{amsfonts}
\usepackage{tabularx}
\usepackage{arydshln}
\usepackage{enumitem}
\usepackage{wrapfig}

\usepackage[resetlabels,labeled]{multibib}
\newcites{A}{Appendix References}

\newcommand{\mypar}[1]{\vspace{0.5pt}\noindent\textbf{#1.}}
\newcommand{\mypartwo}[1]{\vspace{0.5pt}\noindent\textit{#1.}}

\begin{document}
\title{Rethinking Business Process Simulation: \newline A Utility-Based Evaluation Framework}
%
\titlerunning{Rethinking BPS: A Utility-Based Evaluation Framework}
%
\author{Konrad Özdemir\thanks{Equal contribution}\inst{1} \and
Lukas Kirchdorfer$^\star$\inst{1,2} \and \\
Keyvan Amiri Elyasi$^\star$\inst{1}\and
Han van der Aa\inst{3}\and
Heiner Stuckenschmidt\inst{1}
}
\authorrunning{K. Özdemir and L. Kirchdorfer et al.}
%
\institute{Data and Web Science Group, University of Mannheim, Germany\\
\email{\{konrad,keyvan,heiner\}@informatik.uni-mannheim.de} \and
SAP Signavio, Walldorf, Germany\\
\email{lukas.kirchdorfer@sap.com} \and
Faculty of Computer Science, University of Vienna, Austria\\
\email{han.van.der.aa@univie.ac.at}}
\maketitle              
\begin{abstract}
Business process simulation (BPS) is a key tool for analyzing and optimizing organizational workflows, supporting decision-making by estimating the impact of process changes. 
The reliability of such estimates depends on the ability of a BPS model to accurately mimic the process under analysis, making rigorous accuracy evaluation essential. 
However, the state-of-the-art approach to evaluating BPS models has two key limitations. 
First, it treats simulation as a forecasting problem, testing whether models can predict unseen future events.
This fails to assess how well a model captures the as-is process, particularly when process behavior changes from train to test period. Thus, it becomes difficult to determine whether poor results stem from an inaccurate model or the inherent complexity of the data, such as unpredictable drift.
Second, the evaluation approach strongly relies on Earth Mover’s Distance-based metrics, which can obscure temporal patterns and thus yield misleading conclusions about simulation quality.
To address these issues, we propose a novel framework that evaluates simulation quality based on its ability to generate representative process behavior. Instead of comparing simulated logs to future real-world executions, 
we evaluate whether predictive process monitoring models trained on simulated data perform comparably to those trained on real data for downstream analysis tasks.
Empirical results show that our framework not only helps identify sources of discrepancies but also distinguishes between model accuracy and data complexity, offering a more meaningful way to assess BPS quality.

\end{abstract}
%
%
%

\section{Introduction}
Business process simulation (BPS) plays a crucial role in the analysis and redesign of organizational processes. By creating a digital process twin\cite{Dumas21}, simulation enables the estimation of the impact of a process change on key performance indicators, such as cycle time, resource utilization, and waiting time for specific activities---an approach commonly referred to as counterfactual reasoning or ``what-if'' analysis\cite{FundamentalsOfBPM}. By providing such estimates in advance, BPS can significantly enhance decision-making by improving efficiency and reducing the risks associated with process redesign~\cite{FundamentalsOfBPM}. Furthermore, automated approaches that derive process simulation models from historical execution data~\cite{RozinatMSA09,Camargo_2020,MeneghelloFGR25,camargo_DSIM,Kirchdorfer2024} eliminate the need for manual model construction, which is both time-consuming and error-prone~\cite{Aalst15}.
However, the reliability of process simulation hinges on the accuracy of the underlying model. Only models that faithfully replicate the as-is process behavior can provide meaningful and trustworthy insights into the effects of potential changes. This raises a fundamental question: how can we effectively assess the quality of a simulation model? 

In this work, we argue that the current approach to evaluating BPS models has two key issues. 
First, it frames simulation as a forecasting problem, evaluating the quality of a BPS model by comparing simulated event logs to unseen future process executions. 
We argue that this does not evaluate the model’s accuracy in capturing the as-is process, in particular when the process behavior in the test period differs significantly from the training period. Thus, the current approach makes it difficult to determine whether poor results stem from an inaccurate BPS model or from dynamics not captured in the training data, such as an increasing workload due to higher customer demand reflected by more frequent case arrivals during the test period.
Second, the metrics commonly used to compare simulated logs against real logs are mostly based on the Earth Mover's Distance (EMD). However, we give theoretical and empirical evidence that EMD can obscure temporal patterns and has a bias toward favoring 
estimates close to the mean,
potentially yielding misleading conclusions about simulation quality.

In this work, we critically examine the limitations of the current evaluation approach for BPS and propose a novel framework that better aligns with the fundamental purpose of simulation. Rather than treating simulation as a forecasting problem, we advocate for an evaluation paradigm that assesses whether a simulation model generates process behavior that accurately mimics the observed as-is process.
Our proposed framework shifts the focus from direct log-to-log comparisons toward assessing the \emph{utility} of simulated data in downstream tasks. Specifically, we evaluate the quality of simulated event logs by measuring their effectiveness in training predictive process monitoring (PPM) models. If a model trained on simulated data---tasked with predicting next activities, remaining time, or other process-related properties---achieves performance comparable to one trained on real event logs, it suggests that the simulation model provides a realistic representation of reality. This utility-based perspective offers a more meaningful assessment of simulation accuracy and allows to discern between the pure model quality and data complexity. 

The remainder starts with discussing related work in~\autoref{sec:rel_work}, before motivating our new framework by showing the limitations of the existing approach in~\autoref{sec:motivation}. Afterward,~\autoref{sec:approach} describes our evaluation framework with an experimental evaluation in~\autoref{sec:evaluation}. Finally,~\autoref{sec:conclusion} concludes our work.

\section{Related Work} 
\label{sec:rel_work}
This section reviews prior work on BPS model evaluation and discusses the use of downstream tasks for assessing the quality of synthetic data in the broader Machine Learning (ML) community.

\mypar{Evaluation of BPS Models}
The evaluation of BPS models has long been secondary to the development of novel simulation approaches, often treated as a by-product rather than a research focus in itself. 
Early approaches, such as Rozinat et al. \cite{RozinatMSA09}, relied on manual comparisons of simulated and real event logs, analyzing activity execution times and gateway probabilities. Khodyrev and Popova \cite{KhodyrevP14} incorporated the idea of a temporal train-test split, comparing simulated and test logs along metrics such as number of case arrivals and activity durations.
With the rise of data-driven simulation, Camargo et al. \cite{CamargoDR21} proposed a more structured evaluation combining control-flow and temporal aspects, using metrics like control-flow log similarity, mean absolute error (MAE) of cycle times, and EMD of activity durations. In a subsequent work \cite{camargo_DSIM}, the same authors refined this approach, using MAE of cycle times and EMD of activity timestamps, also influencing later BPS studies \cite{meneghello_RIMS}.
Despite these advancements, evaluation in BPS remained fragmented until Chapela-Campa et al. \cite{chapela2025} explicitly tackled the question of how to assess BPS models. Their work introduced a more holistic evaluation framework, incorporating metrics from control-flow, temporal, and congestion perspectives, primarily relying on EMD and Wasserstein-1 distance to compare simulated and test logs. This framework has quickly become the de facto standard, having been applied in several recent BPS studies \cite{Kirchdorfer2024,MeneghelloFGR25,LopezPintadoMD24}. 
However, while this framework represents a significant step forward, it has critical shortcomings, as we will demonstrate in \autoref{sec:motivation}.

\mypar{Downstream Tasks as Means for Evaluation}
Since BPS focuses on the generation of synthetic data, our work draws inspiration from the way in which the quality of synthetically generated data is assessed in other contexts. Specifically, a common practice is the assessment of data quality through downstream task performance, 
 following the \emph{Train on Synthetic, Test on Real} (TSTR) paradigm \cite{TSTR}. This evaluation method has been widely applied across domains, including time series augmentation \cite{timegan}, natural language processing \cite{nlp_augmentation}, object detection \cite{object_detection_TSTR}, and crowd counting \cite{crowd_counting_TSTR}. In TSTR, a predictive model is trained on generated synthetic data and evaluated on hold-out real-world test data, with its performance compared to a model trained on real data. This approach provides a meaningful assessment of the generative model, as it measures how well the generated data captures relevant patterns required for generalization and how much utility has been retained by the generation w.r.t. a specific ML task \cite{TSTR}.
While TSTR has predominantly been used for image and text data, its adaptation to process data presents new opportunities. A key challenge lies in identifying suitable downstream tasks that account for process-specific characteristics such as time-dependent behavior and resource constraints. We close this gap by adapting the TSTR principle to the process mining domain, including appropriate downstream models and tasks.

\section{Motivation}
\label{sec:motivation}
In this section, we examine the current state-of-the-art approach to evaluating BPS models proposed by Chapela-Campa et al.~\cite{chapela2025}, which we refer to as the \textit{Standard Practice}. We then identify and discuss its limitations, which can be summarized by two key issues: \textit{Objective Mismatch} and \textit{Metric Shortfall}. The former addresses the conflation of simulation and forecasting objectives, while the latter provides theoretical and empirical evidence of critical flaws in the evaluation metrics used.


\subsection{The Standard Practice of BPS Evaluation} 
\label{subsec:standard-practice}

\mypar{Preliminaries}
Automated BPS approaches rely on historical process execution data, which is typically captured in \emph{event logs} \cite{FundamentalsOfBPM}.
An event log $\mathcal{L}$ is a multi-set of traces, where each trace $\sigma \in \mathcal{L}$ represents a sequence of events $(e_1, \dots, e_n)$, capturing the execution of a single process instance. Each event $e_i$ is defined as a tuple $\langle a, r, \tau_{\text{start}}, \tau_{\text{end}}\rangle$, where $a$ denotes the executed activity, $r$ the responsible resource, and $\tau_{\text{start}}$ and $\tau_{\text{end}}$ the corresponding start and end timestamps.


\mypar{Evaluation Procedure} The state-of-the-art approach for evaluating simulation accuracy follows a structured flow consisting of five key steps and is largely inspired by the works of Chapela-Campa et al. (cf. \cite{chapela2025}).

\begin{enumerate}[noitemsep, topsep=0pt]
    \item The traces of a given event log $\mathcal{L}$ are partitioned temporally into a training $\mathcal{L}_{\mathrm{train}}$ and a testing log $\mathcal{L}_{\mathrm{test}}$.
    
    \item A BPS model $S$ is discovered using $\mathcal{L}_{\mathrm{train}}$. For example, via \textit{Simod} \cite{Camargo_2020}, \textit{AgentSimulator} \cite{Kirchdorfer2024}, or \textit{DeepSimulator} \cite{camargo_DSIM}.
    
    \item  $S$ is used to simulate a new event log $\mathcal{L}_{\mathrm{sim}}$, starting at the same time as $\mathcal{L}_{\mathrm{test}}$ and including the same number of traces.

    \item To facilitate a direct comparison between $\mathcal{L}_{\mathrm{sim}}$ and $\mathcal{L}_{\mathrm{test}}$, \textit{proxies}\footnote{We use the term proxy in the econometric sense, referring to a variable that serves as a representation of another variable of interest that cannot be directly quantified.}
    reflecting the property of interest are derived from each event log. The prevalent approach for proxy derivation across nearly all metrics involves sequence binning, where events are grouped into $B$ one-hour intervals based on their $\tau_{\text{start}}$ and $\tau_{\text{end}}$ timestamps. Counting the events in each bin produces two \textit{count-sequences} $\bar{x} = (x_1, \dots, x_B)$ for $\mathcal{L}_{\mathrm{sim}}$ and $\bar{y} = (y_1, \dots, y_B)$ for $\mathcal{L}_{\mathrm{test}}$. \label{step:motivation-count-sequences}
    
    \item The discrepancy between $\mathcal{L}_{\mathrm{sim}}$ and $\mathcal{L}_{\mathrm{test}}$ is quantified using the Wasserstein-1 ($W_{1}$) distribution distance applied to the binned \textit{count-sequences}. Formally: $\textrm{Dist}(\mathcal{L}_{\mathrm{sim}}, \mathcal{L}_{\mathrm{test}}) = W_1(\bar{x},\bar{y})$. Simply put, $W_1$ measures the \textit{minimal effort} needed to redistribute probability mass from one histogram to match the other. In this sense, a lower score indicates a better result. Following Chapela-Campa et al. (cf. \cite{chapela2025}), we refer to the $W_1$ distance only, as it poses a more efficient Earth Mover's Distance implementation.\footnote{For more details: \url{https://github.com/konradoezdemir/Rethinking-BPS}}
\end{enumerate}

\noindent Having established the \textit{Standard Practice}, we now highlight the most critical issue inherent in this approach.

\subsection{Objective Mismatch} 
\label{sec:Objective Mismatch}
Establishing a simulation model involves extracting meaningful properties from a reference dataset ($\mathcal{L}_{\mathrm{train}}$) and embedding these into the model $S$ to effectively serve as a digital process twin \cite{Dumas21}. The overall quality of this model should then be measured by how well the model's simulated dataset ($\mathcal{L}_{\mathrm{sim}}$) captures the reference dataset's statistical properties~\cite{endres2022synthetic}. The \textit{Standard Practice} violates this principle through an \textit{Objective Mismatch}: the simulated event log is compared to a test log that may significantly diverge from the training log. In our view, this conflates simulation with forecasting and undermines a clear assessment of model quality. Consequently, we propose to fix this mismatch by using $\mathcal{L}_{\mathrm{train}}$ instead of $\mathcal{L}_{\mathrm{test}}$ as reference for assessing $\mathcal{L}_{\mathrm{sim}}$. The following scenario illustrates how this mismatch can distort evaluation outcomes.

\mypar{Scenario: Time-Varying Behavior in Standard Practice} \label{ex:Time-Varying Behavior in Standard Practice}  
In \autoref{tab:results_drift_chapela}, we examine a \textit{Loan Application} process (cf. \cite{chapela2025}) under two scenarios: the \emph{original} log with stable arrival rates and another where we introduced a \emph{drift} (arrival rate surge) close to the start of the test period. 
First, we simulate two event logs via \textit{Agentsimulator}~\cite{Kirchdorfer2024}, one for the \textit{original}- and one for the \textit{drift}-scenario. 
Second, six distribution distances were used to evaluate each simulated log against its respective test log: \textit{NGD} (N-Gram), \textit{AEDD} (Absolute Event), \textit{CADD} (Case Arrival), \textit{CEDD} (Circadian Event), \textit{REDD} (Relative Event), and \textit{CTDD} (Cycle Time)~\cite{chapela2025}. Now, the results obtained in the \emph{drift} scenario might lead one to conclude that the simulation model is rather poor, as indicated by high error rates in \autoref{tab:results_drift_chapela}. However, \textit{Agentsimulator} actually provides a good representation of the process, as indicated by small error values in the \emph{original} scenario. In fact, it is the evaluation method under the \textit{Standard Practice} that does not adequately distinguish between the intrinsic quality of the BPS model and the complexity introduced by concept drift. For example, \textit{AEDD} increases from 2.97 under stable conditions to 95.03 when a drift is introduced, while \textit{CADD} rises from 0.00 to 121.87. In conclusion, this scenario demonstrates that the perceived quality of a BPS model can appear extremely poor when evaluated using the \textit{Standard Practice} on a process with changing behavior.

\begin{table}[htbp]
    \centering
    \setlength{\tabcolsep}{10pt} 
    \caption{Comparison of distribution distances between the \emph{Original} version of the Loan Application process and a version with altered arrival rate (\emph{Drift}).}
    \label{tab:results_drift_chapela}
    \resizebox{\textwidth}{!}{%
    \begin{tabular}{lrrrrrr}
        \toprule
        Log & NGD & AEDD & CADD & CEDD & REDD & CTDD \\
        \midrule
        Original & 0.07 & 2.97 & 0.00 & 0.27 & 1.66 & 2.71 \\
        Drift & 0.20 & 95.03 & 121.87 & 0.49 & 26.32 & 32.85 \\
        \bottomrule
    \end{tabular}%
    }
\end{table}

\noindent Process behavior changes, as in the above scenario, are not merely superficial. In fact, half of the processes commonly examined in BPS studies \cite{camargo_DSIM,MeneghelloFGR25,Kirchdorfer2024} exhibit substantial drifts in cycle time (cf. \autoref{tab:datasets}). 

One might naturally think that 
this \textit{Objective Mismatch} issue can be resolved by computing distribution distances between the simulated log $\mathcal{L}_{\mathrm{sim}}$ and the training log $\mathcal{L}_{\mathrm{train}}$, rather than the test log $\mathcal{L}_{\mathrm{test}}$.
However, we argue that even when this mismatch is accounted for, the \textit{Standard Practice} remains inadequate, as we discuss next.

\subsection{Metric Shortfall}
From a statistical perspective, $W_1$-based distance metrics (e.g., \textit{AEDD}, \textit{CADD}) require large sample sizes to accurately estimate the underlying probability distributions. This ensures that the computed differences between these estimates are both robust and reliable. In high-dimensional settings where the sample complexity can grow significantly, this necessity becomes especially pronounced (cf.~\cite{wasserstein_curse_of_dim}). Considering this, we encounter a fundamental difficulty when applying the $W_1$ distance to said \textit{count-sequences} (cf. step~\ref{step:motivation-count-sequences}): How should these sequences be interpreted in light of the assumptions made by $W_1$-based metrics? One option is to view the entire sequence as a single representation of the event log. However, this leads to a drastic sampling bias because each log ($\mathcal{L}_{\mathrm{sim}}$ and $\mathcal{L}_{\mathrm{test}}$) is represented by only one sequence. Alternatively, and as implicitly assumed in these distance metrics, one might treat each individual count within the sequence as an \textit{independent observation} from the underlying event log’s distribution. While this approach aligns with the formulation of the $W_1$ distance, it neglects the temporal dependencies and structural relationships between activities that are inherent to event logs.

We formally prove that---despite the statistical appeal of treating individual counts as independent realizations---this approach undermines the proper representation of temporal properties, as detailed in the following theorem.


\begin{theorem} \label{thm_w1}
Let $\mathbf{P}$ and $\mathbf{Q}$ denote the probability distributions associated with $\mathcal{L}_{\mathrm{train}}$, $\mathcal{L}_{\mathrm{test}}$ with samples $(x_i)_{i=1}^B$ and $(y_i)_{i=1}^B$ in $\mathbb{R}^1$. Then, one can show\footnote{For a formal proof: \url{https://github.com/konradoezdemir/Rethinking-BPS}} that the Wasserstein-1 distance admits the following form:
\begin{equation}
W_1(\mathbf{P}, \mathbf{Q}) 
\;=\; 
\frac{1}{B} \sum_{i=1}^B \bigl| x_{(i)} - y_{(i)} \bigr|.\label{PropEMDeq}
\end{equation}
\end{theorem}

As established in \autoref{thm_w1}, computing the $W_{1}$ distance between two \textit{count-sequences} requires sorting each sequence in non-decreasing order; indicated via `$(i)$'. In doing so, the smallest value from the first sequence is compared with the smallest from the second, the second smallest with the second smallest, etc. This reordering removes any original temporal structure, rendering likely inter-dependencies between events irrelevant. A simple example may look as follows:

\begin{example}\label{ex:thm_w1}
Consider hourly call center dial-ins from 10am to 3pm (i.e., $B=5$ hours), which are down-trending with true counts $y_{\cdot}=(5,4,3,1,1)$ ($\mathbf{Q}$). Consider a pattern-neglecting \textit{bad} estimate $\tilde{x}_{\cdot}=(1,3,1,5,4)$ ($\widetilde{\mathbf{P}}$), and a pattern-recognizing \textit{good} estimate $x_{\cdot}=(5,5,3,1,1)$ ($\mathbf{P}$). Sorting non-decreasingly yields $y_{(\cdot)}=(1,1,3,4,5)$, $\tilde{x}_{(\cdot)}=(1,1,3,4,5)$ and $x_{(\cdot)}=(1,2,3,5,5)$. By \autoref{thm_w1}, $W_1(\widetilde{\mathbf{P}},\mathbf{Q})=0$ and $W_1(\mathbf{P},\mathbf{Q})=\frac{1}{5}\left(|4-5|\right)=0.2,$ preferring $\tilde{x}_{\cdot}$ over $x_{\cdot}$ here.
\end{example}

This example underscores that, by neglecting temporal ordering, the $W_1$ distance can falsely favor a bad estimate of the true data ($\tilde{x}_{\cdot}$), while a more realistic one ($x_{\cdot}$) is penalized. Moreover, \autoref{thm_w1} reveals another important aspect about $W_1$: \textit{susceptibility to outliers}. With similarity to the \textit{MAE}, once a largely deviating pair of values occurs, the overall distance falls at risk to explode. With the following scenario, we illustrate how the \textit{Standard Practice} can fail in selecting the optimal simulation model under this exact premise.

\begin{wrapfigure}{r}{0.5\textwidth} 
    \centering
    \includegraphics[width=\linewidth]{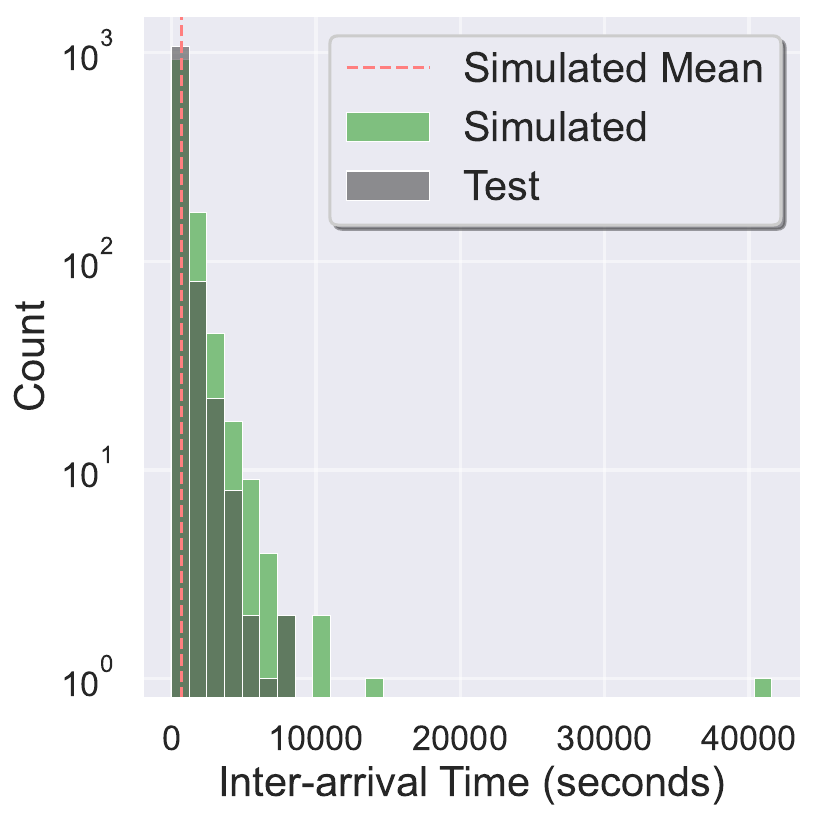}
    \caption{Inter-arrival time histograms for the BPIC12W log.}
    \label{fig:inter-arrival}
\end{wrapfigure}

\mypar{Scenario: Model Selection via Standard Practice}
\autoref{fig:inter-arrival} overlays three histograms representing empirical inter-arrival time distributions for the \emph{BPIC12W} event log. One stems from the test log $\mathcal{L}_{\textrm{test}}$, while the others are generated by \textit{AgentSimulator} \cite{Kirchdorfer2024}. In the `Simulated' case, \textit{AgentSimulator} discovers and models arrivals (i.e., $\mathcal{L}_{\textrm{sim}}$), whereas in `Simulated Mean', arrivals are generated using only the mean inter-arrival time from the train log $\mathcal{L}_{\textrm{train}}$. While both `Simulated'
  and `Test' exhibit similar inter-arrival distributions (resembling an exponential distribution), `Simulated Mean' collapses into a degenerate distribution, concentrating all mass at a single value. From a practitioner's perspective, the choice is clear: the mean-based approach is unsuitable for meaningful simulation, whereas \textit{AgentSimulator} reasonably approximates the distribution associated with $\mathcal{L}_{\textrm{test}}$. However, under the \textit{Standard Practice}, i.e., using the \textit{CADD} metric, `Simulated Mean' paradoxically achieves a better score (42.7) than `Simulated' (55.8). 
This is because, as outlined before, the $W_{1}$ distance is susceptible to outliers. This means that it heavily \textit{penalizes} large, wrong estimates and instead, tends to favor observations close to the mean value (cf.~\cite{villani2003topics}). This phenomenon is clearly illustrated in the histogram: although `Simulated' effectively covers the `Test' distribution better than `Simulated Mean', the outlier estimate just after the 40,000 mark produces a substantial error that significantly increases the overall score. Consequently, when model selection is based on this metric, one would inadvertently favor a model that employs the simulated mean approach, even though it fails to capture the true variability of the event log.

In light of these pitfalls, we propose a new, comprehensive evaluation framework for business process simulation.

\section{Utility-Based Evaluation Framework}
\label{sec:approach}
\begin{figure}[htbp]
    \centering
    \includegraphics[trim = 0mm 0mm 0mm 0mm, clip, width=\columnwidth]{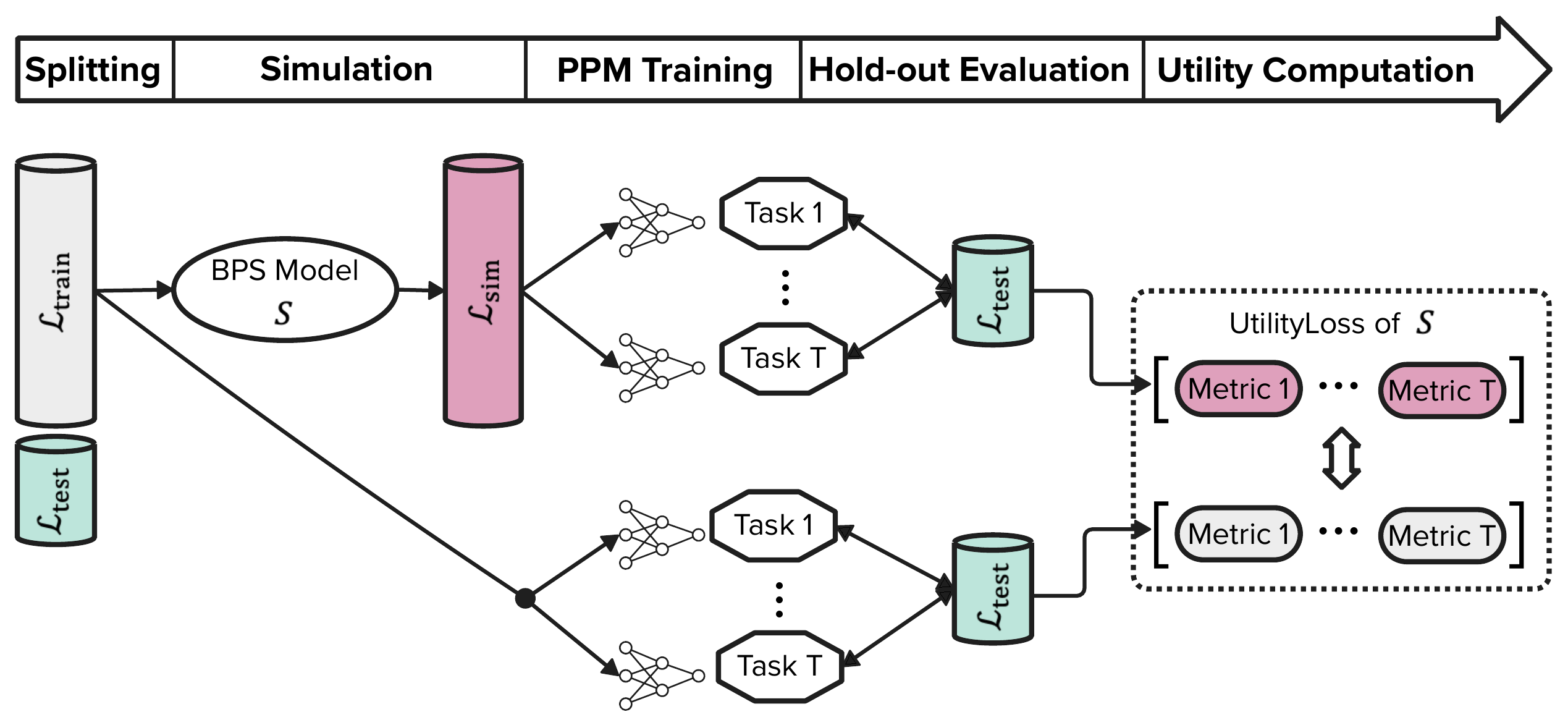}
    \caption{Overview of our proposed BPS evaluation framework.}
    \label{fig:eval_schematic}
\end{figure}

In this section, we introduce the details of our utility-based BPS evaluation framework. The core idea is to assess the quality of BPS models by measuring how well their simulated data supports downstream prediction tasks compared to real training data. As shown in the visualization in \autoref{fig:eval_schematic}, our proposed evaluation framework consists of five steps. We describe each of these next, while also highlighting key differences to the \textit{Standard Practice} discussed in \autoref{subsec:standard-practice}.

\mypar{Splitting} 
The first step is to split an event log temporally into train log $\mathcal{L}_{\textrm{train}}$ and test log $\mathcal{L}_{\textrm{test}}$.

\mypar{Simulation}
Then, $\mathcal{L}_{\textrm{train}}$ is used to train a BPS model \textit{S}. Unlike the \textit{Standard Practice}, which aims to align $\mathcal{L}_{\textrm{sim}}$ with the characteristics of $\mathcal{L}_{\textrm{test}}$---implicitly treating simulation as a forecasting task and leading to an objective mismatch---we ensure that \textit{S} generates a simulated log $\mathcal{L}_{\textrm{sim}}$ that mirrors the properties of $\mathcal{L}_{\textrm{train}}$. Specifically, $\mathcal{L}_{\textrm{sim}}$ should contain the same number of cases as $\mathcal{L}_{\textrm{train}}$ and start at the same point in time.

\mypar{PPM Training} 
Next, we train predictive process monitoring (PPM) models w.r.t. $T$ different \emph{downstream} tasks, separately using real data $\mathcal{L}_{\textrm{train}}$ and simulated data $\mathcal{L}_{\textrm{sim}}$. This results in two distinct predictive models per task $t \in \{1,\dots,T\}$: $\textrm{PM}_t(\mathcal{L}_{\textrm{train}})$ and $\textrm{PM}_t(\mathcal{L}_{\textrm{sim}})$.
To enable a comprehensive evaluation, our framework incorporates prediction tasks spanning the four key process perspectives commonly assessed in BPS evaluation~\cite{chapela2025}:

\mypartwo{Control-Flow} The control-flow perspective captures the order and dependency relations among activities. To capture this, we instruct our models to predict the next activity (NAP) in ongoing cases, assessing whether the BPS model preserves event order.

\mypartwo{Resource} The resource perspective captures the workforce of the process. By predicting the next role (NRP) in ongoing cases, we assess whether the BPS model considers role assignment rules, ordering, and interaction patterns.

\mypartwo{Temporal} The temporal perspective captures process timing, including event durations, case durations, and inter-event intervals. This is covered by predicting the next activity processing time (NPP), the next waiting time between the end of the previous activity and the start of the new activity (NWP), and the remaining time until the completion of the case (RTP).

\mypartwo{Congestion} The congestion perspective captures the workload over time, including queuing effects and resource contention. We cover this perspective by already introduced tasks: NWP (waiting time), RTP (remaining time), and NRP (role). These tasks are suitable proxies for congestion as they capture the amount of waiting time (e.g., due to resource contention), the overall length that a process instance stays in the system, and role assignments.

Note that the downstream tasks described above may not encompass all relevant process characteristics, such as comprehensive workload or queuing dynamics. However, the modular structure of our framework makes it easy to extend the set of tasks to include additional process characteristics as needed.
Also, it is the practitioner's choice if one model should be trained per task or if all tasks should be trained in a single model (Multi-Task Learning). Additionally, to account for potential model bias, each task can be trained using $K$ different architectures---such as an LSTM and a Transformer model, as will be done in our evaluation in \autoref{sec:evaluation}.

\mypar{Hold-out Evaluation} 
Having trained the PPM models on both real data $\mathcal{L}_{\textrm{train}}$ and simulated data $\mathcal{L}_{\textrm{sim}}$, we assess their performance on the hold-out test log $\mathcal{L}_{\textrm{test}}$. Each task $t \in \{1,\dots,T\}$ is evaluated using a specific metric $\mathcal{M}_t$, such as \emph{MAE} for RTP and accuracy for NAP.
Rather than combining these metrics into a single score, we maintain a \emph{process perspective-specific evaluation} to preserve interpretability. Thus, for an event log $\mathcal{L} \in \{{\mathcal{L}_{\textrm{train}}, \mathcal{L}_{\textrm{sim}}\}}$, the resulting metrics for $T$ tasks---averaged over $K$ model architectures---are represented as a vector:

\begin{equation}
    \mathcal{M}(\mathcal{L}) = \left[ \frac{1}{K} \sum_{k=1}^{K} \mathcal{M}^{\mathcal{L}_{\textrm{test}}}_1(\textrm{PM}_1^k(\mathcal{L})), \dots, \frac{1}{K} \sum_{k=1}^{K}\mathcal{M}^{\mathcal{L}_{\textrm{test}}}_T(\textrm{PM}_T^k(\mathcal{L})) \right]
\end{equation}



\mypar{Utility Computation}
To evaluate how well a BPS model $S$ preserves the predictive utility of the original process data, we introduce the concept of \emph{utility loss}. This metric quantifies the extent to which a model trained on simulated data deviates in performance from one trained on real data across various downstream tasks. The rationale behind this choice is that task-specific performance metrics can serve as a proxy for the practical utility of the data. Computing these metrics for both the real and simulated logs yields $\mathcal{M}(\mathcal{L}_{\textrm{train}})$ and $\mathcal{M}(\mathcal{L}_{\textrm{sim}})$. These vectors can then be compared element-wise to assess how much utility is lost (or preserved) when training PPM models on simulated instead of real data. A small absolute deviation in accuracy or error suggests that the BPS model \textit{S} successfully captures key patterns of $\mathcal{L}_{\textrm{train}}$, while a larger deviation---regardless of direction---indicates a loss in fidelity. Formally, we define the utility loss of a model \textit{S} as the element-wise absolute difference between $\mathcal{M}(\mathcal{L}_{\textrm{train}})$ and $\mathcal{M}(\mathcal{L}_{\textrm{sim}})$:

\begin{equation}
    \textrm{Utility}{\textrm{Loss}}(S) = \big| \mathcal{M}(\mathcal{L}_{\textrm{train}}) - \mathcal{M}(\mathcal{L}_{\textrm{sim}}) \big|.
\end{equation}

\noindent Note that our framework is intended to be used for the evaluation of BPS models capable of supporting what-if analysis. Therefore, although we recognize that approaches that effectively copy the initial event log $\mathcal{L}_{\textrm{train}}$ would yield a near-optimal UtilityLoss, they fail to meet a fundamental requirement of BPS. As such, they are not suitable for evaluation within our framework.

\section{Evaluation}
\label{sec:evaluation}
This section outlines the experiments conducted to evaluate our utility-based framework for measuring BPS quality. We perform two experiments: The first (\autoref{sec_evaluation_1}) tests the framework's applicability by verifying whether known modifications to a simulation model are reflected accordingly. The second (\autoref{sec_evaluation_2}) benchmarks state-of-the-art BPS approaches, highlighting their strengths, weaknesses, and insights gained through our framework. Our Python implementation and additional results are available in our public repository (cf. \cite{codebase_rethinking_zenodo}).

\subsection{Experiment 1: Applicability of Our Framework}
\label{sec_evaluation_1}
To validate our framework's applicability, we examine whether known modifications to a BPS model result in a corresponding UtilityLoss. An effective evaluation should (i) identify inaccurate models and (ii) reveal which perspective of the process they fail to capture appropriately.
We first describe the experimental setup before discussing the results.

\mypar{Experimental Setup}
In this experiment, we largely follow the synthetic evaluation from Chapela-Campa et al.~\cite{chapela2025}, using the \textit{Loan Application} process as the basis (cf.~\cite{chapela2025}; see~\autoref{tab:datasets} for details). We temporally split the event log trace-wise into 80\% training and 20\% test data. The process comprises 12 activities, beginning with \emph{Check application form completeness}. Its control-flow structure includes a loop, a parallel branch involving three activities, three exclusive gateways, and three possible end points: \emph{Approve application}, \emph{Reject application}, and \emph{Cancel application}. In total, the process is carried out by 19 distinct resources.


\mypartwo{Scenarios} 
Given the train set of this process, we use the \textit{Simod} BPS approach~\cite{Camargo_2020} to discover a ground truth simulation model (Loan$_{GT}$) and create different modifications (following Chapela-Campa et al.~\cite{chapela2025}), which will allow us to assess whether our framework effectively penalizes these models in the respective downstream tasks. For each of the following models\footnote{The specifics of these modifications and their respective logs are in our repository.}, we simulate 10 event logs that align with the train log in terms of start time and number of cases:
\begin{itemize}[noitemsep, topsep=0pt]
    \item Loan$_{SEQ}$: arranging the three parallel activities as a sequence. 
    \item Loan$_{S{\text -}G}$: altering, on top of Loan$_{SEQ}$, the branching probabilities.
    \item Loan$_{RC}$: halving the available resources. 
    \item Loan$_{EXT}$: adding extraneous waiting time to delay the start of activities. 
    \item Loan$_{DUR}$: increasing the duration of the activities of the process.
    \item Loan$_{CAL}$: changing resource working schedules from 9am-5pm to 2pm-10pm.
    \item Loan$_{ARR}$: increasing the rate of case arrivals.
\end{itemize}

\mypartwo{Tasks and evaluation metrics}
For downstream utility assessment, we consider the five PPM tasks
described in~\autoref{sec:approach}. To quantify loss, we use \textit{accuracy} as the metric for the two classification tasks (NAP and NRP) and \textit{MAE} for the three regression tasks (NPP, NWP, and RTP).

\mypartwo{PPM approach} 
We use \emph{ProcessTransformer}~\cite{bukhsh2021processtransformer} as the approach for the PPM tasks.
Since this model was designed to just predict the next activity (NAP), remaining time (RTP), and the next timestamp, we adapted it so that it can distinguish between processing and waiting times, thus enabling the NPP and NWP tasks. We also extended it to predict the role responsible for the next activity, enabling NRP.
We use the model's default configuration.

\mypar{Results and Discussion} \autoref{tab:loanapp_modifications} presents the predictive performance obtained with the simulated data from the ground truth model Loan$_{GT}$ and the UtilityLoss for each modification, illustrating how our framework effectively detects alterations in the respective process perspectives.

\begin{table}[htbp]
    \centering
    \setlength{\tabcolsep}{3pt} 
    \caption{Average utility loss (with standard deviation) for modifications of the Loan$_{GT}$ process. NPP, NWP, and RTP errors are measured in minutes.}
    \label{tab:loanapp_modifications}
    \begin{tabular}{lrrrrr}
        \toprule
          & NAP & NRP & NPP & NWP & RTP \\
        \midrule
        Loan$_{GT}$       & 0.71 (0.01)      & 0.75 (0.00)       & 67.60 (2.80)          & 12.13 (1.48)         & 238.26 (8.38)    \\
        \noalign{\vskip 1mm}
        \cdashline{1-6}
        \noalign{\vskip 1mm}
        Loan$_{SEQ}$                & 0.26 (0.03)      & 0.26 (0.04)       & 16.12 (3.29)          & 12.03 (0.52)         & 35.88 (18.04)    \\
        Loan$_{S{\text -}G}$           & 0.32 (0.07)      & 0.35 (0.04)       & 12.40 (2.91)          & 715.84 (112.23)        & 4131.58 (1802.63)   \\
        Loan$_{RC}$       & 0.00 (0.01)      & 0.12 (0.14)       & 16.55 (3.31)          & 164.28 (63.69)        & 771.89 (368.17)   \\
        Loan$_{EXT}$         & 0.00 (0.01)      & 0.00 (0.00)       & 24.30 (5.19)          & 74.37 (6.92         & 508.74 (53.15)    \\
        Loan$_{DUR}$           & 0.00 (0.01)       & 0.00 (0.00)       & 49.63 (7.42)         & 7.27 (6.01)         & 263.29 (39.86)    \\
        Loan$_{CAL}$          & 0.00 (0.01)       & 0.00 (0.00)       & 0.44 (5.38)          & 0.45 (1.68)         & 5.43 (8.93)    \\
        Loan$_{ARR}$           & 0.00 (0.01)       & 0.00 (0.00)       & 10.60 (2.28)          & 35.14 (3.14)         & 90.1 (15.21)    \\
        \bottomrule
    \end{tabular}
\end{table}

\mypartwo{Control-flow}
Significant deviations from the ground truth model (Loan$_{GT}$) in the control-flow perspective are expected only in cases where the activity order is altered, as in Loan$_{SEQ}$ and Loan$_{S{\text -}G}$. Since these changes disrupt the original sequence of events, our framework correctly assigns a UtilityLoss in NAP for these two models, while all other models align with the ground truth, as expected.

\mypartwo{Resource}
The resource-related metric NRP should reflect deviations not only for Loan$_{SEQ}$ and Loan$_{S{\text -}G}$ (where activity order changes can impact resource allocation) but also for Loan$_{RC}$, where resource assignments are altered due to halving available resources. Our framework appropriately penalizes these models, capturing the expected disruptions in resource allocation.

\mypartwo{Temporal}
In the temporal perspective, the most substantial UtilityLoss occurs---as expected---in NPP for Loan$_{DUR}$, where activity durations are explicitly modified. Since duration changes influence process timing, this effect propagates to RTP, demonstrating our framework’s ability to capture temporal deviations.

\mypartwo{Congestion}
Regarding congestion, NWP shows notable losses for Loan$_{S{\text -}G}$, Loan$_{RC}$, Loan$_{EXT}$, and Loan$_{ARR}$ aligning with the expected impact of sequential execution, resource contention, extraneous delays, and higher arrival frequency respectively. These congestion effects propagate to RTP, reflecting significantly longer cycle times for the three mentioned modifications.

Notably, Loan$_{CAL}$, modifying only absolute timestamps without affecting the relative temporal structure, correctly receives a near-zero UtilityLoss, as our framework evaluates temporal relationships rather than absolute timestamps.

In summary, our framework effectively identifies BPS models that deviate from the ground truth and precisely determines which process perspective is inadequately captured, ensuring a comprehensive evaluation of BPS accuracy.

\subsection{Experiment 2: Benchmark}
\label{sec_evaluation_2}
In this experiment, we showcase the practicality of our framework in benchmarking state-of-the-art BPS approaches. We first explain the experimental setup and then discuss the results.

\mypar{Experimental Setup} For the benchmark, we rely on 8 event logs\footnote{Datasets can be downloaded from here: \url{https://zenodo.org/records/5734443}} (see details in \autoref{tab:datasets}) commonly used to evaluate BPS approaches~\cite{Kirchdorfer2024,MeneghelloFGR25,camargo_DSIM}. As in the first experiment, we split event logs temporally, with 80\% for training and 20\% for testing. Each BPS model generates 10 simulated logs per process, which are used to train and evaluate the PPM models. We aggregate results across these 10 runs and compare them to models trained on real data. To account for performance variance, PPM models trained on real data use 10 random seeds. The tasks and evaluation metrics in Experiment 1 (\autoref{sec_evaluation_1}) are also applied here.

\begin{table*}[htbp]
\centering
\setlength\tabcolsep{7pt} 
\caption{Description of event log properties. Average cycle time for train and test sets (CT-train, CT-test) are reported in days.}
\label{tab:datasets}
\begin{tabular}{lrrcrr}
\toprule
\textbf{Log} & \textbf{Traces} & \textbf{Events} & \textbf{\#Act/\#Res} & \textbf{CT-train} & \textbf{CT-test}\\ 
\midrule
 Loan App & 1000 & 7492 & 12 / 19 & 0.42 & 0.46 \\ 
 P2P & 608 & 9119 & 21 / 27 &  12.14 & 30.82\\ 
 C. 1000 & 1000 & 38160 & 42 / 14 & 0.96 & 0.80\\ 
 C. 2000 & 2000 & 77418 & 42 / 14 & 0.86 & 0.77 \\ 
 CVS & 10000 & 103906 & 15 / 6 & 5.74 & 12.05 \\
 Production & 225 & 4503 & 24 / 41 & 17.44 & 4.32\\ 
 CDM & 954 & 6870 & 18 / 432 & 10.54 & 3.60\\  
 BPI12W & 8616 & 59302 & 6 / 52 & 7.92 & 8.16 \\ 
 BPI17W & 30276 & 240854 & 8 / 136 & 12.47 & 11.50 \\ 

 \hline
\end{tabular}
\end{table*}

\mypartwo{BPS approaches}
We evaluate three state-of-the-art BPS approaches: 

\begin{itemize}[noitemsep, topsep=0pt]
    \item \textit{Simod} \cite{Camargo_2020} is a traditional approach that derives a BPMN model along with a set of simulation parameters. In this work, we use an enhanced version of \textit{Simod} that accounts for differentiated resource behavior~\cite{Lopez-PintadoD22}. 

    \item \textit{DeepSimulator (DSim)}~\cite{camargo_DSIM} is a hybrid approach that integrates a BPMN model to govern control flow while leveraging multiple deep learning models for learning temporal dynamics.

    \item \textit{AgentSimulator (ASim)}~\cite{Kirchdorfer2024} is an agent-based approach that represents each resource as an autonomous agent, simulating the process through agent interactions.
    
\end{itemize}

\mypartwo{PPM approaches} We use two PPM model architectures in our benchmark evaluation. The first, \emph{ProcessTransformer}~\cite{bukhsh2021processtransformer}, is described in \autoref{sec_evaluation_1}. The second, proposed by Camargo et al.~\cite{camargo_DGEN}, represents an \emph{LSTM} architecture, capable of predicting the five downstream tasks. For each task, we train an LSTM with two hidden layers of size 50, each, and use fixed-size n-grams of 10.

\begin{table}[t!]
    \centering
    \setlength{\tabcolsep}{4pt}
    \caption{Utility-based performance comparison of BPS models. }
    \begin{tabular}{ccrrrrr}
        \toprule
        Log & Data & NAP & NRP & NPP(min) & NWP(hour) & RTP(day) \\
        \midrule
        \multirow{4}{*}{\rotatebox{90}{P2P}} & real & 0.84 (0.02) & 0.86 (0.01) & 67.11 (9.38) & 62.88 (0.76) & 17.82 (0.63)\\
        \noalign{\vskip 1mm}
        \cdashline{2-7}
        \noalign{\vskip 1mm}
        & Simod & 0.03 (0.00) & 0.36 (0.00) & 65.13 (0.00) & \textbf{3.44} (0.00) & \textbf{1.44} (0.00) \\
        & DSim & 0.14 (0.02) & NA & 51.46 (51.7) & 6.12 (4.54) & 6.65 (9.81) \\
        & ASim & \textbf{0.00} (0.02) & \textbf{0.35} (0.02) & \textbf{0.60} (5.37) & 9.74 (2.56) & 3.05 (3.92)\\
        \midrule 
        \multirow{4}{*}{\rotatebox{90}{C.1000}} & real & 0.72 (0.01) & 0.44 (0.01) & 19.14 (0.15) & 0.28 (0.01) & 0.39 (0.03) \\
        \noalign{\vskip 1mm}
        \cdashline{2-7}
        \noalign{\vskip 1mm}
        & Simod & 0.22 (0.02) & \textbf{0.06} (0.02) & 2.83 (0.40) & 2.07 (0.39) &  3.53 (1.48)\\
        & DSim & 0.52 (0.04) & NA & 6.44 (2.84) & \textbf{0.19} (0.04) & \textbf{0.07 (0.06)}\\
        & ASim & \textbf{0.13} (0.02) & 0.09 (0.01) & \textbf{0.20} (0.05) & 2.83 (0.40) & 0.17 (0.07)\\
        \midrule 
        \multirow{4}{*}{\rotatebox{90}{C.2000}} & real & 0.73 (0.01) & 0.46 (0.01) & 19.30 (0.29) & 0.25 (0.00) & 0.34 (0.03) \\
        \noalign{\vskip 1mm}
        \cdashline{2-7}
        \noalign{\vskip 1mm}
        & Simod & 0.20 (0.01) & \textbf{0.07} (0.02) & 2.83 (0.73) & 1.90 (0.38) &  2.58 (1.38)\\
        & DSim & 0.53 (0.03) & NA & 5.75 (2.19) & 0.34 (0.08) & 0.86 (1.37) \\
        & ASim & \textbf{0.10} (0.02) & 0.08 (0.07) & \textbf{2.40} (0.27) & \textbf{0.15} (0.04) & \textbf{0.09} (0.04)\\
        \midrule 
        \multirow{4}{*}{\rotatebox{90}{CVS}} & real & 0.76 (0.01) & 0.80 (0.01) & 3.26 (0.10) & 20.31 (0.25) & 3.16 (0.05)\\
        \noalign{\vskip 1mm}
        \cdashline{2-7}
        \noalign{\vskip 1mm}
        & Simod & 0.27 (0.01) & 0.60 (0.04) & 0.28 (0.20) & 14.64 (1.79) & 15.09 (1.05) \\
        & DSim & 0.14 (0.02) & NA & 0.29 (0.22) & 31.83 (9.75) & 9.88 (2.00) \\
        & ASim & \textbf{0.05 }(0.01) & \textbf{0.02} (0.01) & \textbf{0.09} (0.01) & \textbf{14.17} (2.44) & \textbf{0.54} (0.03)\\
        \midrule 
        \multirow{4}{*}{\rotatebox{90}{Production}} & real & 0.57 (0.02) & 0.49 (0.03) & 120.83 (1.91) & 12.45 (0.99) &  8.18 (2.11)\\
        \noalign{\vskip 1mm}
        \cdashline{2-7}
        \noalign{\vskip 1mm}
        & Simod & 0.33 (0.02) & 0.22 (0.02) & 34.74 (32.17) & 27.11 (10.90) & \textbf{1.83} (12.34) \\
        & DSim & 0.46 (0.06) & NA & \textbf{12.20} (6.74) & \textbf{8.25} (1.38) & 91.06 (58.85) \\
        & ASim & \textbf{0.02} (0.02) & \textbf{0.10} (0.06) & 16.68 (4.02) & 15.88 (1.53) & 3.89 (0.74)\\
        \midrule 
        \multirow{4}{*}{\rotatebox{90}{CDM}} & real & 0.74 (0.01) & 0.62 (0.04) & 6.90 (0.48) & 10.45 (0.49) &  2.91 (0.07)\\
        \noalign{\vskip 1mm}
        \cdashline{2-7}
        \noalign{\vskip 1mm}
        & Simod & 0.21 (0.05) & \textbf{0.29} (0.08) & 1.25 (0.96) & 13.60 (2.25) & 5.39 (0.40) \\
        & DSim & 0.45 (0.07) & NA & \textbf{0.13} (0.50) & 5.22 (0.76) &  \textbf{0.32} (0.14)\\
        & ASim & \textbf{0.15} (0.03) & 0.34 (0.11) & 0.22 (1.06) & \textbf{3.03} (0.15) & 0.72 (0.02)\\
        \midrule 
        \multirow{4}{*}{\rotatebox{90}{BPI12W}} & real & 0.67 (0.03) & 0.80 (0.00) & 8.81 (0.16) & 28.06 (0.22) & 5.62 (0.52) \\
        \noalign{\vskip 1mm}
        \cdashline{2-7}
        \noalign{\vskip 1mm}
        & Simod & 0.43 (0.00) & 0.01 (0.00) & \textbf{0.26} (0.00) & 8.77 (0.00) & 0.42 (0.00) \\
        & DSim & 0.44 (0.02) & NA & 0.90 (0.38) & 2.25 (0.76) & 1.13 (0.20) \\
        & ASim & \textbf{0.01} (0.01) & \textbf{0.00} (0.00) & 0.27 (0.09) & \textbf{0.19} (0.16) & \textbf{0.20} (0.91)\\
        \midrule 
        \multirow{4}{*}{\rotatebox{90}{BPI17W}} & real & 0.67 (0.03) & 0.94 (0.00) & 6.12 (0.41) & 33.12 (0.29) & 6.20 (0.23) \\
        \noalign{\vskip 1mm}
        \cdashline{2-7}
        \noalign{\vskip 1mm}
        & Simod & 0.43 (0.00) & 0.17 (0.00) & \textbf{0.40} (0.00) & 4.68 (0.00) & \textbf{0.69} (0.00) \\
        & DSim & 0.44 (0.02)& NA & 1.77 (1.94) & 10.93 (6.03) & 4.35 (1.44) \\
        & ASim & \textbf{0.01} (0.01) & \textbf{0.08} (0.07) & 0.82 (1.29) & \textbf{4.61} (3.67) & 1.67 (2.32)\\
        \bottomrule    
    \end{tabular}
    \label{tab:real_res_agg}
\end{table}

\mypar{Results and Discussion}
\autoref{tab:real_res_agg} summarizes the results\footnote{More detailed results on individual architecture runs can be found in our repository.} of our second experiment. Each log presents results obtained using real training data, followed by the UtilityLoss of each BPS approach, separated by a dashed horizontal line. The results shown are averaged over both PPM model architectures, with the best-performing BPS approach per task boldened. 

Overall, no single BPS approach consistently dominates across all logs and tasks. However, ASim consistently outperforms others in the control-flow dimension (NAP), while also most frequently leading in resource (NRP) and temporal tasks (NPP, NWP). Regarding RTP, results are rather mixed, with both \textit{Simod} and ASim leading in three logs, respectively. In contrast, DSim only claims two. 

ASim’s robust performance in control-flow accuracy is attributed to its reliance on observed frequentist activity transition probabilities. Conversely, \textit{Simod} and DSim employ model discovery algorithms with pruning, simplifying the discovered model but compromising accuracy in simulated activity sequences.

Unlike the aggregated temporal perspective offered by the \textit{Standard Practice}, our evaluation separately addresses next activity processing (NPP) and waiting times (NWP). Notably, NPP and NWP exhibit no correlation, reinforcing the importance of their independent assessment to reveal distinct BPS improvement ideas.
While \textit{Simod} and ASim model processing times using parameterized distributions, and account for extraneous delays, DSim relies on LSTM models. Given DSim’s poorer results for temporal metrics, the necessity of computationally expensive, black-box deep learning approaches in BPS warrants reconsideration.

Our evaluation framework's ability to separate model accuracy from data complexity is particularly evident in the P2P log, which exhibits substantial cycle time drift between the training and test sets (\autoref{tab:datasets}). PPM approaches trained on real data show significantly higher errors across all temporal tasks compared to other event logs. While the \textit{Standard Practice} would simply suggest poor BPS model performance, our framework provides deeper insights. For example, although ASim's absolute error in the NPP task is high, its UtilityLoss is nearly zero, indicating that ASim effectively captures processing times despite the dataset's inherent complexity due to evolving process characteristics.

Finally, our framework facilitates diagnosing root causes behind poor BPS performance. In the CVS log, for instance, \textit{Simod} and ASim yield similar results for NPP and NWP, yet \textit{Simod} exhibits an exceptionally high RTP UtilityLoss (15.09 days), far exceeding ASim’s 0.54 days. Given their comparable temporal metrics, this suggests that \textit{Simod}’s poor RTP performance primarily stems from inaccuracies in its control-flow and resource modeling (NAP and NRP). This highlights the interdependence of process perspectives and underscores the necessity of ensuring accuracy across all simulation components, as errors in one aspect can propagate and severely degrade overall simulation utility.

In summary, the benchmark shows that our framework can effectively compare BPS approaches and identify their strengths and weaknesses across different process perspectives.


\section{Conclusion}
\label{sec:conclusion}
In this work, we introduced a framework for evaluating the quality of business process simulation (BPS) models by assessing the utility of simulated event logs via downstream tasks from predictive process monitoring (PPM). Unlike traditional evaluation methods that rely on direct log-to-log comparisons in a forecasting manner, our framework shifts the focus towards assessing the \emph{utility} of simulated data in downstream tasks compared to the original data. Our results demonstrate that the proposed downstream tasks capture and differentiate modifications to a process model across control-flow, resource, temporal, and congestion dimensions. 
Additionally, we showed that our framework can effectively benchmark various BPS approaches to identify the respective strengths and weaknesses. Thereby, contrary to previous evaluation methods, it (1) can discern model accuracy from data complexity, (2) focuses on temporal relationships rather than absolute timestamps, and (3) can provide more fine-grained insights for potential areas of model improvement.

\mypar{Limitations}
Despite these advantages, our framework has some key limitations. The current set of downstream tasks, while effective, may not fully capture all relevant process characteristics, presenting an opportunity for further refinement. For instance, the resource perspective so far only considers role assignments, without explicit insights into overall occupation and work in progress. Additionally, our framework introduces significantly higher computational costs compared to traditional evaluation methods, as it requires training predictive models rather than computing log distances. Finally, despite using multiple network architectures and seeds, our evaluation framework may introduce variability due to training stochasticity, potentially limiting reproducibility and comparability. 
However, these limitations are expected, given that our approach introduces a fundamentally new perspective on evaluating BPS models. 

\mypar{Future Work}
Therefore, future work will focus on expanding the set of downstream tasks, exploring alternative predictive model architectures beyond the LSTM and Transformer considered in this study, and improving the efficiency of the framework. Furthermore, the integration of BPS and PPM opens a promising research direction, where simulated data could be leveraged as an augmentation tool to enhance PPM models without requiring additional real-world data.

\section{Acknowledgement}
The authors acknowledge support by the state of Baden-Württemberg through bwHPC.

%
%

\bibliographystyle{splncs04}
\bibliography{bibliography}

\end{document}